\documentclass{article}
\usepackage{spconf}
\usepackage{pbox}
\usepackage{times}
\usepackage{latexsym}
\usepackage{color}
\usepackage{times}
\usepackage{bbm}
\usepackage{dsfont}
\usepackage{qtree}
\usepackage{multirow}
\usepackage{url}
\usepackage{latexsym}
\usepackage{graphicx}
\usepackage{amsmath,amssymb,amsthm}

\newcommand{\R}{\ensuremath{\mathbb{R}}}

\newcommand{\ra}{\ensuremath{\rightarrow}}

\newcommand{\myset}[1]{\left\{#1\right\}}
\newcommand{\paren}[1]{\left(#1\right)}

\newcommand{\abs}[1]{\left|#1\right|}

\newcommand{\by}{\ensuremath{\times}}

\usepackage{url}

\title{\emph{OneNet}: \\Joint Domain, Intent, Slot Prediction for Spoken Language Understanding}

\name{Young-Bum Kim\textsuperscript{*}, Sungjin Lee\textsuperscript{**}, Karl Stratos{***}}
\address{Amazon Alexa Brain, Seattle, WA\textsuperscript{*}, \\
Microsoft Research, Redmond, WA\textsuperscript{**},\\
Toyota Technological Institute, Chicago, IL\textsuperscript{***}
}

\begin{document}

\maketitle

\begin{abstract}
In practice, most spoken language understanding systems process user input in a pipelined manner; first domain is predicted, then intent and semantic slots are inferred according to the semantic frames of the predicted domain.
The pipeline approach, however, has some disadvantages: error propagation and lack of information sharing.
To address these issues, we present a unified neural network that jointly performs domain, intent, and slot predictions. Our approach adopts a principled architecture for multitask learning to fold in the state-of-the-art models for each task. 
With a few more ingredients, e.g. orthography-sensitive input encoding and curriculum training, our model delivered significant improvements in all three tasks across all domains over strong baselines, including one using oracle prediction for domain detection, on real user data of a commercial personal assistant.
\end{abstract}

\begin{keywords}
Natural language understanding, Multitask learning, Joint modeling
\end{keywords}

\section{Introduction}

\begin{figure}[t!]
  \begin{center}
    \includegraphics[width=0.5\textwidth]{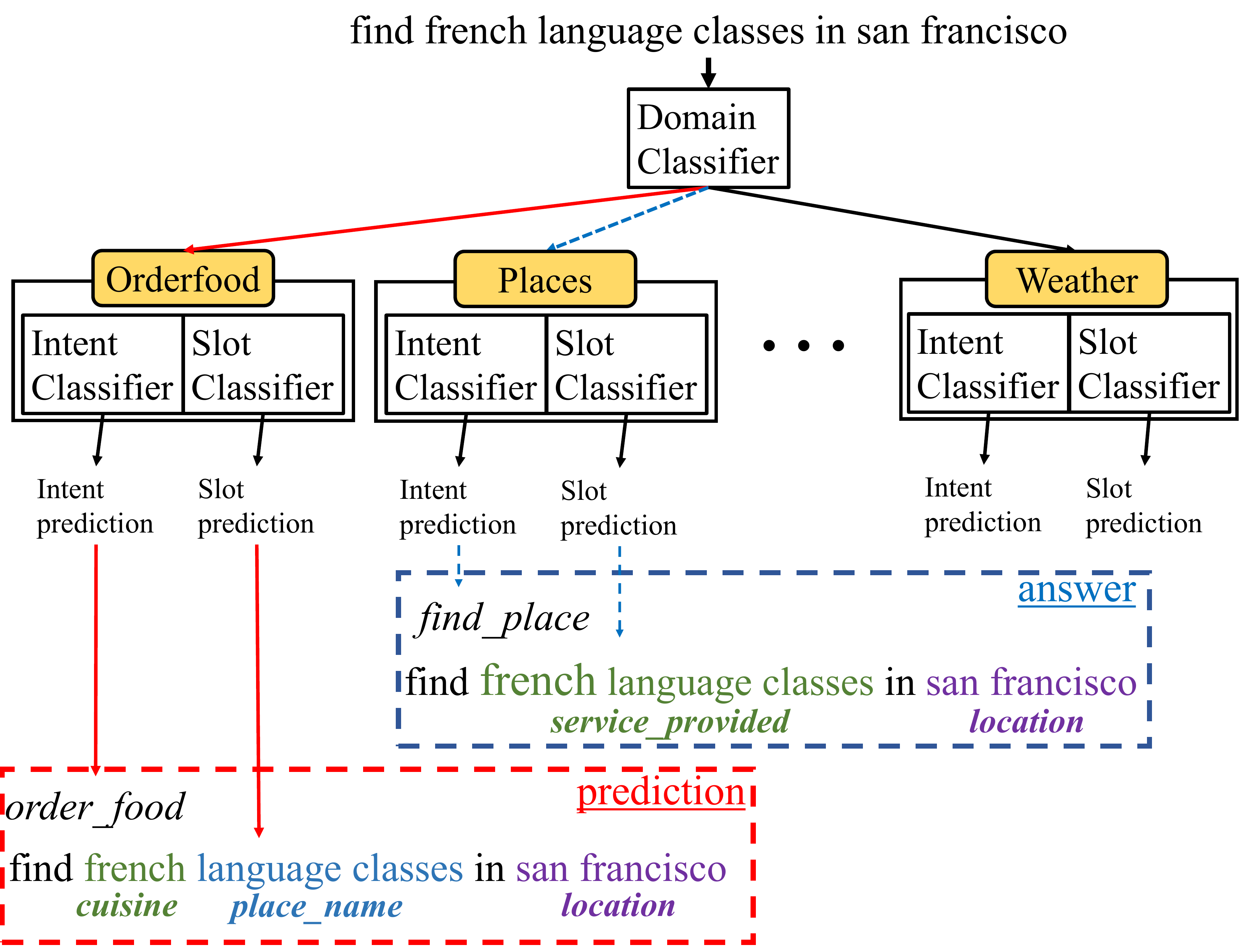}
  \end{center}
  \caption{\small Illustration of the pipelined procedure. If domain classifier predicted a wrong domain \texttt{Orderfood}, subsequent tasks -- intent detection and slot tagging -- would also make errors consequently.}
  \label{fig:pipeline}
\end{figure}

Major personal assistants, e.g., Amazon's Alexa, Apple Siri and Microsoft Cortana, support task-oriented dialog for multiple domains. 
A typical way to handle multiple domains for the spoken language understanding (SLU) task~\cite{sarikaya2016overview} is to perform domain prediction first and then carry out intent prediction~\cite{kim2016scalable,kim2017da,kim2017advr} and slot tagging~\cite{ybkim2015weakly,kimgazet2015,anastasakos2014task, kimpre2015,kim2015new,kim2017pre, kim2017da, kim2017advr} (Figure~\ref{fig:pipeline}). However, this approach has critical disadvantages. First, the error made in domain prediction propagates to downstream tasks - intent prediction and slot tagging. Second, the domain prediction task cannot benefit from the downstream prediction results. Third, it is hard to share domain-invariant features such as common or similar intents and slots across different domains.

Despite such disadvantages, there were a few reasons that such a pipelined approach was widely adopted before the deep neural networks (DNNs) era has come. For instance, in graphical models, jointly modeling multiple tasks often entails an unwieldy model with a high computational complexity~\cite{jeong2008triangular}. Also, with a discrete representation, a unified model that covers all domains usually results in severe data sparseness problems given the relatively small amount of labeled data. Last but not least, one can use domain-specific resources, e.g. domain gazetteers, that often results in a significant performance increase. Domain-specific resources, however, are costly to build in new domains. 

With recent advances, DNNs provide various affordances that allow us to address most issues described above: Multitask learning becomes as easy as dropping in additional loss terms~\cite{collobert2008unified}. Continuous representation learning addresses the data sparseness problem~\cite{bengio2003neural}. Unsupervised learning helps us tap into unlimited data sources, obviating demands for domain-specific resources~\cite{mikolov2013efficient}.

In this paper, we present a unified neural architecture, \emph{OneNet}, that performs domain, intent and slot predictions all together. Our approach adopts a principled architecture for multitask learning to fold in the state-of-the-art models for each task. 
With a few more ingredients, e.g. orthography-sensitive input encoding and curriculum training, our model delivered significant improvements in all three tasks across all domains over strong baselines, including one using oracle prediction for domain detection, on real user data of a commercial personal assistant.

\section{Related Work}
There has been an extensive line of prior studies for jointly modeling intent and slot predictions: a triangular conditional random field (CRF)~\cite{jeong2008triangular}, a convolutional neural networks-based triangular CRF~\cite{xu2013convolutional}, recursive neural networks learning hierarchical representations of input text~\cite{guo2014joint}, several variants of recurrent neural networks~\cite{zhang2016joint,Liu+2016,liu-lane:2016:SIGDIAL}. 

There are also a line of prior works on multi-domain learning to leverage existing domains: 
\cite{ybkim2016reuse} proposed a  constrained  decoding  method  with  a  universal  slot  tagging  model. \cite{jaech2016domain} proposed $K$ domain-specific feedforward layers with a shared word-level Long Short-Term Memory (LSTM)~\cite{hochreiter1997long} layer across domains; \cite{kimfrustratingly} instead employed $K+1$ LSTMs. Finally, there is a few approaches performing multi-domain/multitask altogether with a single network. \cite{kimdomainless}, however, requires a separate schema prediction to construct decoding constraints. \cite{hakkani2016multi} proposed to employ a sequence-to-sequence model by introducing a fictitious symbol at the end of an utterance of which tag represents the corresponding domain and intent. In comparison to the prior models, not only does our approach afford a more straightforward integration of multiple tasks but also it delivers a higher performance. 

\begin{figure*}[!ht]
  \begin{center}
    \includegraphics[width=0.755\textwidth]{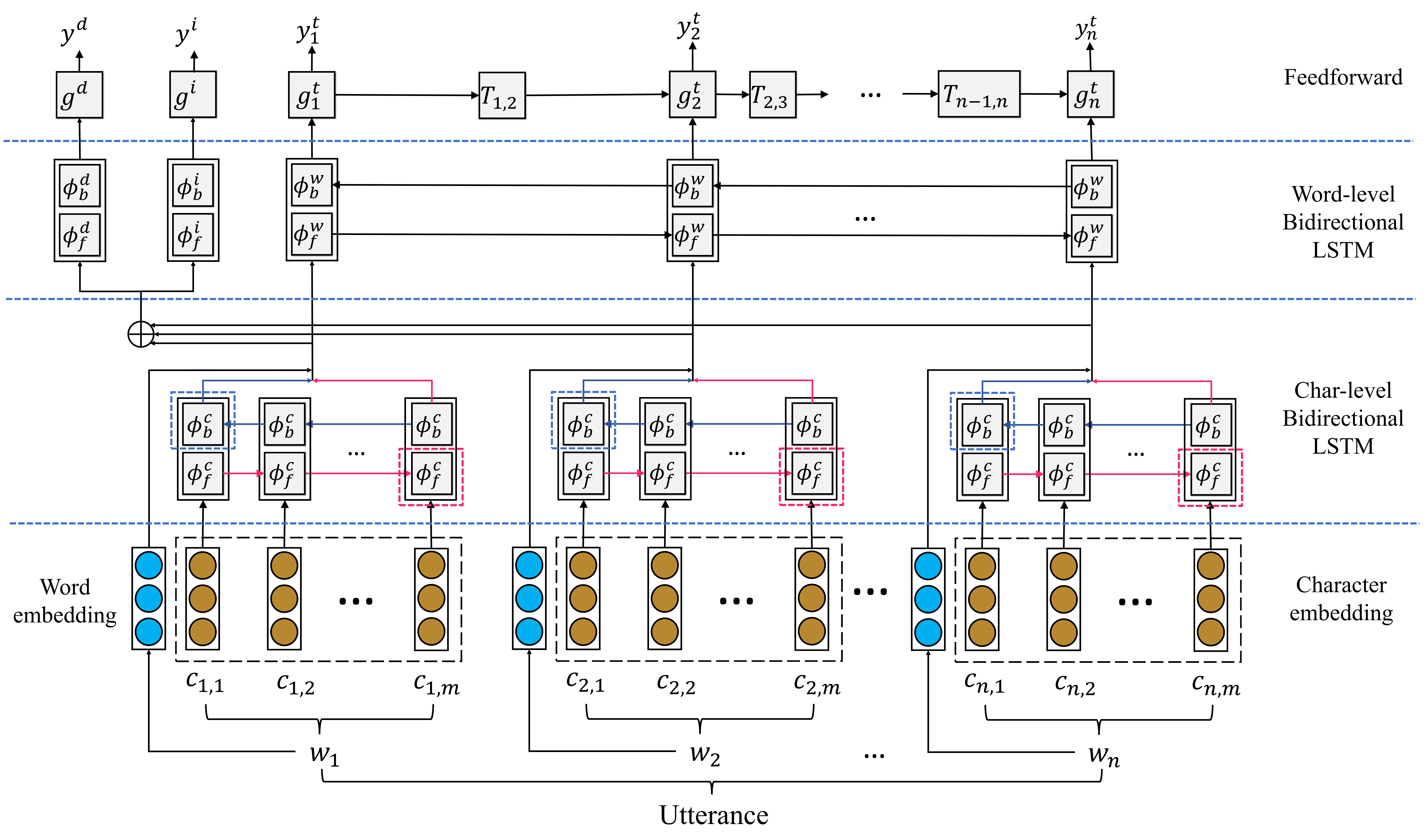}
  \end{center}
  \caption{\small The overall network architecture for joint modeling.}
  \label{modelarchitecture}
\end{figure*}


\section{OneNet}
\label{sec:method}

At a high level, our model consists of builds on three ingredients (Figure~\ref{modelarchitecture}): a shared orthography-sensitive word embedding layer; a shared bidirectional LSTM (BiLSTM) layer that induces contextual vector representations for the words in a given utterance; three separate output layers performing domain, intent and slot prediction, respectively, which are rooted in the shared BiLSTM layer. We combine the losses of these output layers which will be minimized jointly. Crucially, the joint optimization updates the shared layers to be simultaneously suitable for all three tasks,
allowing us to not only avoid the error propagation problem but also improve the performance of individual tasks by sharing task-invariant information.
\subsection{Embedding layer}
In order to capture character-level patterns, we construct a orthography-sensitive word embedding layer following \cite{lample2016neural}.
Let $\mathcal{C}$ denote the set of characters and $\mathcal{W}$ the set of words.
Let $\oplus$ denote the vector concatenation operation.
We use an LSTM simply as a mapping $\phi:\R^d \times \R^{d'} \ra \R^{d'}$
that takes an input vector $x$ and a state vector $h$ to output a new state vector $h' = \phi(x, h)$.
The model parameters associated with this layer are:
\begin{align*}
\textbf{Char embedding: } &e_c \in \R^{25} \text{ for each } c \in \mathcal{C} \\
\textbf{Char LSTMs: } &\phi^{\mathcal{C}}_f, \phi^{\mathcal{C}}_b: \R^{25} \times \R^{25} \ra \R^{25} \\
\textbf{Word embedding: } &e_w \in \R^{100} \text{ for each } w \in \mathcal{W}
\end{align*}
Let $w_1 \ldots w_n \in \mathcal{W}$ denote a word sequence where word $w_i$ has character $w_i(j) \in \mathcal{C}$ at position $j$.
This layer computes a orthography-sensitive word representation $v_i \in \R^{150}$ as
\begin{align*}
f^{\mathcal{C}}_j &= \phi^{\mathcal{C}}_f\paren{e_{w_i(j)}, f^{\mathcal{C}}_{j-1}} &&\forall j = 1 \ldots \abs{w_i} \\
b^{\mathcal{C}}_j &= \phi^{\mathcal{C}}_b\paren{e_{w_i(j)}, b^{\mathcal{C}}_{j+1}} &&\forall j = \abs{w_i} \ldots 1\\
v_i &= f^{\mathcal{C}}_{\abs{w_i}} \oplus b^{\mathcal{C}}_1 \oplus e_{w_i} &&
\end{align*}

\subsection{BiLSTM layer}
A widely successful architecture for encoding a sentence $(w_1 \ldots w_n) \in \mathcal{W}^n$
is given by BiLSTMs \cite{schuster1997bidirectional,graves2012neural}: 
\begin{align*}
\textbf{Word LSTMs: } &\phi^{\mathcal{W}}_f, \phi^{\mathcal{W}}_b: \R^{150} \times \R^{100} \ra \R^{100}\\ 
\end{align*}
for each $i = 1 \ldots n$.\footnote{For simplicity, we assume some random initial
state vectors such as $f^{\mathcal{C}}_0$ and $b^{\mathcal{C}}_{\abs{w_i}+1}$ when we describe LSTMs.}
Next, the model computes

\begin{align*}
f^{\mathcal{W}}_i &= \phi^{\mathcal{W}}_f\paren{v_i, f^{\mathcal{W}}_{i-1}}  &&\forall i = 1 \ldots n \\
b^{\mathcal{W}}_i &= \phi^{\mathcal{W}}_b\paren{v_i, b^{\mathcal{W}}_{i+1}} &&\forall i = n \ldots 1
\end{align*}
and induces a context-sensitive word representation $h_i \in \R^{200}$ as
\begin{align}
h_i &= f^{\mathcal{W}}_i \oplus b^{\mathcal{W}}_i \label{eq:h}
\end{align}

for each $i = 1 \ldots n$. These vectors are used to define the domain/intent classification loss and the slot tagging loss below.
Note that the model parameters associated with both shared layers are collectively denoted as $\Theta$.
\subsection{Domain classification}
We predict the domain of an utterance using $(h_1 \ldots h_n) \in \R^{200}$ in \eqref{eq:h} as follows.
Let $\mathcal{D}$ denote the set of domain types.
We introduce a single-layer feedforward network $g^{{\tiny d}}: \R^{200} \ra \R^{\abs{\mathcal{D}}}$
whose parameters are denoted by $\Theta^{{\tiny d}}$. We compute a $\abs{\mathcal{D}}$-dimensional vector
\begin{align*}
\mu^{{\tiny d}} = g\paren{\sum_{i=1}^n h_i}
\end{align*}
and define the conditional probability of the correct domain $\tau$ as
\begin{align}
p(\tau | h_1 \ldots h_n) \propto \exp\paren{\mu^{{\tiny d}}_{\tau}} \label{eq:domain}
\end{align}
The domain classification loss is given by the negative log likelihood:
\begin{align}
L^{{\tiny d}}\paren{\Theta, \Theta^{{\tiny d}}} = - \sum_l \log p\paren{\tau^{(l)} | h^{(l)}} \label{eq:domain-loss}
\end{align}
where $l$ iterates over the domain-annotated utterances.

\subsection{Intent classification}

The intent classification loss is given in an identical manner to the domain classification loss:
\begin{align}
L^{{\tiny i}}\paren{\Theta, \Theta^{{\tiny i}}} = - \sum_l \log p\paren{\tau^{(l)} | h^{(l)}} \label{eq:intent-loss}
\end{align}
where $l$ iterates over the intent-annotated utterances.

\begin{table*}[!ht]
\small
\centering
\begin{tabular}{c|c|c||c|c|c}
System & Type                                                                     & Model                           & Domain           & Intent           & Slot           \\ \hline
1      & \multirow{2}{*}{\shortstack{Domain adapt\\Independent mode}}               & \cite{jaech2016domain}       & 91.06\%          & 87.05\%          & 85.80          \\
2      &                                                                          & \cite{kimfrustratingly}      & 91.06\%          & 89.24\%          & 88.17          \\   \hline 
3      & \multirow{5}{*}{\shortstack{Independent domain\\Joint intent and slot}}  & \cite{xu2013convolutional}   & 91.06\%          & 86.83\%          & 85.50          \\
4      &                                                                          & \cite{guo2014joint}          & 91.06\%          & 89.46\%          & 88.00          \\
5      &                                                                          & \cite{zhang2016joint}        & 91.06\%          & 89.99\%          & 87.69          \\
6      &                                                                          & \cite{Liu+2016}              & 91.06\%          & 90.75\%          & 88.47          \\
7      &                                                                          & \cite{liu-lane:2016:SIGDIAL} & 91.06\%          & 90.84\%          & 88.39          \\ \hline
8      & \multirow{2}{*}{Full joint}                                              & \cite{hakkani2016multi}      & 92.05\%          & 89.56\%          & 87.52          \\
9      &                                                                          & \cite{kimdomainless}         & -                & 90.11\%          & 88.46          \\ \hline
10     & Independent models                                                       & \emph{OneNet-Independent}       & 91.06\%          & 86.47\%          & 85.50          \\ \hline
11     & Oracle domain                                                            & \emph{OneNet-Oracle}            & 100.0\%          & 93.98\%          & 90.83          \\ \hline
12     & Full joint                                                               & \emph{OneNet}                   & \textbf{95.50}\% & \textbf{94.59}\% & \textbf{93.20} \\ \hline
\end{tabular}
\caption{\small Comparative results of our joint model against other baseline systems. System 9 does not make domain prediction.}
\label{tab:predicted_domain_result}
\end{table*}

\subsection{Slot tagging loss}
Finally, we predict the semantic slots of the utterance using $(h_1 \ldots h_n) \in \R^{200}$ in \eqref{eq:h} as follows.
Let $\mathcal{E}$ denote the set of semantic types and $\mathcal{L}$ the set of corresponding BIO label types,
that is, $\mathcal{L} = \myset{\texttt{B-}e: e \in \mathcal{E}} \cup \myset{\texttt{I-}e: e \in \mathcal{E}} \cup \myset{\texttt{O}}$.
We add a transition matrix $T \in \R^{\abs{\mathcal{L}} \by \abs{\mathcal{L}}}$ and a single-layer feedforward network $g^{{\tiny t}}: \R^{200} \ra \R^{\abs{\mathcal{L}}}$
to the network; denote these additional parameters by $\Theta^t$.
The CRF tagging layer defines a joint distribution over label sequences of $y_1 \ldots y_n \in \mathcal{L}$ of $w_1 \ldots w_n$ as

\begin{align}
  p(y_1 \ldots& y_n |  h_1 \ldots h_n) \notag\\
  &\propto \exp\paren{\sum_{i=1}^n T_{y_{i-1}, y_i} \times g^t_{y_i}(h_i)} \label{eq:crf}
\end{align}
The tagging loss is given by the negative log likelihood:
\begin{align}
L^{{\tiny t}}\paren{\Theta, \Theta^t} = - \sum_l \log p\paren{y^{(l)} | h^{(l)}} \label{eq:global-loss}
\end{align}
where $l$ iterates over tagged sentences in the data.

\subsection{Joint loss}
The final training objective is to minimize the sum of the domain loss \eqref{eq:domain-loss}, intent loss \eqref{eq:intent-loss},
and tagging loss \eqref{eq:global-loss}:
\begin{align}
L(\Theta, \Theta^d, \Theta^i, \Theta^t) = \sum_{\alpha \in \{\Theta^d,\Theta^i,\Theta^t\}}
 L^\alpha \paren{\Theta, \Theta^\alpha} \label{eq:joint-loss} \nonumber
\end{align}
In stochastic gradient descent (SGD), this amounts to computing each loss $l^d, l^i, l^t$ separately at each annotated utterance and then taking a gradient step on $l^d + l^i + l^t$.
Observe that the shared layer parameters $\Theta$ are optimized for all tasks.

\subsection{Curriculum learning}
We find in experiments that it is important to pre-train individual models.
Specifically, we first optimize the domain classifier \eqref{eq:domain-loss},
then the intent classifier \eqref{eq:intent-loss},
then jointly optimize the domain and intent classifiers \eqref{eq:domain-loss} + \eqref{eq:intent-loss},
and finally jointly optimize all losses \eqref{eq:domain-loss} + \eqref{eq:intent-loss} + \eqref{eq:global-loss}.
Separately training these models in the beginning makes it easier for the final model to improve upon individual models.

\section{Experiments}
\subsection{Data}
The data is collected from 5 domains of Microsoft Cortana, a commercial personal assistant: \emph{alarm}, \emph{calendar}, \emph{communication}, \emph{places}, \emph{reminder}. Please refer to Appendix~\ref{sec:data} for detailed data statistics. The numbers of utterances
for training, tuning, and test are 10K, 1K, and 15K respectively for all domains.

\subsection{Training setting}
\label{subsec:setting}
All models were implemented using Dynet \cite{neubig2017dynet} and were trained using Adam \cite{kingma2014adam}\footnote{learning rate=$4 \times 10^{-4}$, $\beta_1=0.9$, $\beta_2=0.999$, and $\epsilon=1e-8$.}. Each update was computed with Intel Math Kernel Library\footnote{https://software.intel.com/en-us/articles/intelr-mkl-and-c-template-libraries} without minibatching. We used the dropout regularization~\cite{srivastava2014dropout} with the keep probability of 0.4. We used pre-trained embeddings used by \cite{lample2016neural}.

\subsection{Results} 
To evaluate the \emph{OneNet} model, we conducted comparative experiments with a rich set of prior works (Table~\ref{tab:predicted_domain_result}) on real user data. We report domain, intent classification results in accuracy and slot tagging results in slot F1-score. 
For those systems which do not include the domain prediction task (marked as Indep. Domain and Pipeline for the Type category), we used the part for the domain prediction of the \emph{OneNet} model trained only with the domain loss. 
The result clearly shows that our joint approach enjoys a higher performance than various groups of baseline systems: ones performing domain adaptation for three independent prediction models (1-2); ones exercising joint prediction only for intent detection and slot tagging (3-7); finally the full joint models (8-9). Particularly, the large performance gap between \emph{OneNet-Independent} and \emph{OneNet} demonstrates how much gain we could get through the proposed joint modeling. Interestingly, \emph{OneNet} even outperformed \emph{OneNet-Oracle} that does not incorporate the domain prediction part in training, instead referring to the oracle domain labels. This result displays an unintuitive power of shared multitask representation learning at a first glance. We found character-level modeling generally adds about 1.5\% on average performance, better capturing sub-word patterns, eg. prefix, suffix, word shape, and making the model robust to spelling errors.
Also, without curriculum learning, we saw about 1.7\% performance drop in intent prediction. By placing intent prediction earlier in training, the latent representation is geared more toward intent prediction.
A detailed performance breakdown for the \emph{OneNet} variants by domain are provided in Appendix~\ref{sec:breakdown}. 
Furthermore, an illustrative example where our joint model get the prediction right while non-joint models do incorrectly can be found in Appendix~\ref{sec:example}.


\section{Conclusion}
To address the disadvantages of the widely adopted pipelined architecture for most SLU systems, we presented a unified neural network, \emph{OneNet}, that jointly performs domain, intent and slot predictions. Our model delivered significant improvements in all three tasks across all domains over strong baselines, including one using oracle prediction for domain detection, on real user data of a commercial personal assistant. 
Future work can include an extension of \emph{OneNet} to take into account dialog history aiming for a holistic framework that can handle contextual interpretation as well.

\bibliography{acl2017}
\bibliographystyle{IEEEbib}

\appendix
\section{Data Statistics}
\label{sec:data}

\section{Performance breakdown by domain}
\label{sec:breakdown}
\begin{table*}[t!]
\small
\centering
\begin{tabular}{cccccccccccc}

& \multicolumn{3}{c}{Oracle Domain} & & \multicolumn{3}{c}{Pipeline} & & \multicolumn{3}{c}{Joint} \\ \cline{2-4} \cline{6-8} \cline{10-12} Domain        & Domain & Intent & Slot  & &Domain & Intent & Slot & &  Domain & Intent & Slot \\ \hline
alarm         & 100\%        & 95.07\%      & 93.93       & & 94.10\%      & 91.00\%      & 89.65  & &93.77\%      & 95.92\%      & 94.37 \\     
calendar      & 100\%        & 92.80\%      & 81.81       & & 83.99\%      & 78.83\%      & 74.46  & &96.89\%      & 93.19\%      & 83.12 \\         
comm.         & 100\%        & 89.99\%      & 85.72       & & 92.57\%      & 84.19\%      & 81.52  & &94.41\%      & 90.91\%      & 88.89 \\     
places        & 100\%        & 94.47\%      & 77.30       & & 94.83\%      & 90.31\%      & 75.27  & &95.84\%      & 95.03\%      & 81.36 \\     
reminder      & 100\%        & 97.58\%      & 86.27       & & 89.80\%      & 88.03\%      & 79.21  & &96.61\%      & 97.89\%      & 88.43 \\ \hline    
AVG           & 100\%        & 93.98\%      & 85.01       & & 91.06\%      & 86.47\%      & 80.02  & &95.50\%      & 94.59\%      & 87.23 \\     \hline

\end{tabular}
\caption{Performance breakdown for the \emph{OneNet} variants by domain}
\label{tab:breakdown}
\end{table*}
A detailed performance breakdown for the \emph{OneNet} variants by domain are provided in Table~\ref{tab:breakdown}.

\section{Example prediction}
\label{sec:example}
Table~\ref{tab:example_slot_tags} shows an illustrative example where our joint model get the prediction right while non-joint models do it incorrectly.
\begin{table*}[!ht]
\small
\centering
\begin{tabular}{c|ccc}
Type             & Sentence & Domain & Intent       \\ \hline
Gold             & \pbox{9cm}{inform \underline{maya}\textsubscript{contact name} about \underline{change of lunch tomorrow from eleven} \underline{thirty to twelve}\textsubscript{message}} & comm. & send text             \\ 
\emph{OneNet-Pipeline}  & \pbox{9cm}{inform maya about change of \underline{lunch}\textsubscript{title} \underline{tomorrow}\textsubscript{start date} from \underline{eleven}\textsubscript{start time} \underline{thirty}\textsubscript{start date} to \underline{twelve}\textsubscript{end time}} & cal.      & change cal entry \\ 
\emph{OneNet-Oracle}        & \pbox{9cm}{inform \underline{maya}\textsubscript{contact name} about \underline{change of lunch}\textsubscript{message} \underline{tomorrow}\textsubscript{date} from \underline{eleven thirty}\textsubscript{time} to \underline{twelve}\textsubscript{time}} & comm. & send text             \\ 
\emph{OneNet}      & \pbox{9cm}{inform \underline{maya}\textsubscript{contact name} about \underline{change of lunch tomorrow from eleven} \underline{thirty to twelve}\textsubscript{message}} & comm. & send text             \\
\end{tabular}
\caption{Example predictions by different models. Gold represents the human labels for the sentence.}
\label{tab:example_slot_tags}
\end{table*}

\begin{table*}[!ht]
\small
\centering
\begin{tabular}{c|ccc}
Domain        & average utterance length   & \# slots & average \#slots/utterance  \\ \hline
alarm         & 5.7 & 15 &  2.12         \\ 
calendar      & 5.9 & 42 &  2.02       \\ 
communication & 3.1 & 44 &  1.50        \\ 
places        & 5.62 & 63 & 2.45           \\
reminder      & 6.12 & 34 & 3.07         \\
\end{tabular}
\caption{Data additional stats}
\label{tab:dataadd}
\end{table*}

\begin{table*}[!ht]
\small
\centering
\begin{tabular}{c|ccccc}
              & alarm   & calendar & communication & places & reminder \\ \hline
alarm         &    	15  &    13    &      3        & 2      & 9   \\ 
calendar      &         & 	 42    &     10        & 4      & 26  \\ 
communication & 	    &          &    44         & 10     &  8  \\ 
places        &         &          &               & 63     & 4   \\
reminder      &         &          &               &        & 34 \\
\end{tabular}
\caption{The number of overlapped slots between domains}
\label{tab:overlap}
\end{table*}

\begin{table*}[!ht]
\small
\centering
\begin{tabular}{c|ccccc}
              & alarm   & calendar & communication & places & reminder \\ \hline
alarm         &   3539  &    2253  &      2682     & 2512   & 2766  \\ 
calendar      &         & 	 12563 &     8477      & 8148   & 8054  \\ 
communication & 	    &          &    77385      & 19847  & 16465 \\ 
places        &         &          &               & 57494  & 15614 \\
reminder      &         &          &               &        & 30707 \\
\end{tabular}
\caption{The number of overlapped vocabularies between domains}
\label{tab:vocabularies}
\end{table*}

\end{document}